\documentclass[conference]{IEEEtran}
\IEEEoverridecommandlockouts
\usepackage{newtxmath} 
\usepackage{cite}
\usepackage{afterpage}
\usepackage{dirtree,float}
\usepackage{textcomp}
\usepackage{flushend}
\usepackage{tikz}
\usepackage{url}
\usepackage{subfigure}
\usepackage{microtype}
\usepackage{hyperref}
\usepackage{graphicx}
\usepackage{booktabs}
\usepackage{tabularx} 
 \usepackage{array}

 \newcommand\copyrighttext{%
   \footnotesize “This work has been accepted to the IJCNN conference and will be published in IEEE proceedings. \\
   Copyright may be transferred without notice, after which this version may no longer be accessible.”}

 \newcommand\copyrightnotice{%
     \begin{tikzpicture}[remember picture,overlay]
        \node[anchor=south,yshift=10pt] at (current page.south) {\fbox{\parbox{\dimexpr\textwidth-\fboxsep-\fboxrule\relax}{\copyrighttext}}};
     \end{tikzpicture}%
 }

\begin{document}
\copyrightnotice
\title{SENMap: Multi-objective data-flow mapping and synthesis for hybrid scalable neuromorphic systems}
\author{
 \IEEEauthorblockN{Prithvish V Nembhani\IEEEauthorrefmark{1}\IEEEauthorrefmark{2}\IEEEauthorrefmark{3}, Oliver Rhodes\IEEEauthorrefmark{1}, Guangzhi Tang\IEEEauthorrefmark{4}}
\IEEEauthorblockN{ Alexandra F Dobrita\IEEEauthorrefmark{2}, Yingfu Xu\IEEEauthorrefmark{2}, Kanishkan Vadivel\IEEEauthorrefmark{2}, Kevin Shidqi\IEEEauthorrefmark{2}, Paul Detterer\IEEEauthorrefmark{2}} , 
 \IEEEauthorblockN{Mario Konijnenburg\IEEEauthorrefmark{2}, Gert-Jan van Schaik\IEEEauthorrefmark{2}, Manolis Sifalakis\IEEEauthorrefmark{2}, Zaid Al-Ars\IEEEauthorrefmark{3}, Amirreza Yousefzadeh\IEEEauthorrefmark{5}}
\IEEEauthorblockA{ \textit{\IEEEauthorrefmark{1}University of Manchester-Advanced Processor Technologies, United Kingdom}, \textit{\IEEEauthorrefmark{2}imec-NL, Netherlands}, 
 \\\textit{\IEEEauthorrefmark{3}Delft University of Technology, Netherlands}, \textit{\IEEEauthorrefmark{4}Mastritch University-DACS, Netherlands}, \textit{\IEEEauthorrefmark{5}University of Twente, Netherlands}}}

\clearpage
\maketitle
\begin{abstract} \label{sec:abstract} 
This paper introduces SENMap, a mapping and synthesis tool for scalable, energy-efficient neuromorphic computing architecture frameworks. SENECA is a flexible architectural design optimized for executing edge AI SNN/ANN inference applications efficiently. To speed up the silicon tape-out and chip design for SENECA, an accurate emulator, SENSIM, was designed. While SENSIM supports direct mapping of SNNs on neuromorphic architectures, as the SNN and ANNs grow in size, achieving optimal mapping for objectives like energy, throughput, area, and accuracy becomes challenging. This paper introduces SENMap, flexible mapping software for efficiently mapping large SNN and ANN applications onto adaptable architectures. SENMap considers architectural, pretrained SNN and ANN realistic examples, and event rate-based parameters and is open-sourced along with SENSIM to aid flexible neuromorphic chip design before fabrication. Experimental results show SENMap enables ~40 percent energy improvements for a baseline SENSIM operating in timestep asynchronous mode of operation. SENMap is designed in such a way that it facilitates mapping large spiking neural networks for future modifications as well. 
\end{abstract}

\section{Introduction}\label{sec:intro}
Recent advances in artificial intelligence (AI) have allowed machines to perform complex tasks traditionally reserved for human intelligence. Deep neural networks (DNNs) have achieved remarkable results in image recognition, natural language processing, and reinforcement learning. As AI tasks become more complex, neuroscience insights are increasingly applied to enhance AI systems. Neuromorphic computing, inspired by the human brain, aims to mimic neural principles in AI architectures. Unlike traditional AI, which relies on discrete computations, neuromorphic systems emulate spiking neuron behavior and temporal dynamics, offering advantages such as improved energy efficiency, robustness, and real-time processing.

Neuromorphic computing, a form of bio-inspired computing, replicates the brain’s capabilities for tasks such as audio/video processing and decision-making \cite{mayr_spinnaker_2019, moradi_scalable_2018, benjamin_neurogrid_2014, corradi_gyro_2021, schemmel_accelerated_2020}. Advances include silicon-based implementations and software emulators such as Lava SDK \cite{noauthor_lava_nodate}, PyCARL \cite{balaji_pycarl_2020}, and Nengo \cite{bekolay_nengo_2014}. Spiking Neural Networks (SNNs), central to neuromorphic computing, use spike-based communication to mimic neuronal activity, improving energy efficiency and biological plausibility \cite{maass_networks_1997, moradi_scalable_2018, pei_towards_2019, ma_darwin_2017}. Neuromorphic systems use specialized cores, such as Loihi crossbars or ARM cores on SpiNNaker, to handle neuron dynamics and synaptic weights. Software tools such as sPyNNaker \cite{rhodes_spynnaker_2018}, PyCARL \cite{balaji_pycarl_2020}, SpiNeMap \cite{balaji_mapping_2020}, and DFSynthesizer \cite{song_dfsynthesizer_2021} optimize SNN mapping. However, scalability and adaptability challenges remain. As larger-scale SNN applications develop \cite{oconnor_deep_2016, zheng_going_2020, hu_spiking_2020, he_deep_2015}, addressing these issues is crucial.

To address scalability, accuracy, and adaptability challenges while improving power/energy efficiency and latency, we introduce SENMap. This paper details SENMap, an extension of SENSIM \cite{nembhani_sensim_2024}, which enhances SNN mapping and synthesis, focusing on mapping, design space exploration, and closed-loop synthesis for practical applications on SENECA. SENMap is adaptable to various neuromorphic designs. Section~\ref{sec:SENECA_SENSIM} gives a brief description of SENECA \cite{yousefzadeh_seneca_2022,tang_seneca_2023,tang_open_2023,xu_optimizing_2024} and SENSIM, and Section~\ref{sec:lit_review} discusses parallel developments to SENMap and compares SENMap with other neuromorphic mapping tools. Section~\ref{sec:senmap} describes the design of SENMap and Section~\ref{sec:experiments} presents the details of the experimentation and results. Section~\ref{sec:conclusion_future_work} concludes the paper summarizing contributions and future work.

\section{SENECA \& SENSIM BRIEF}\label{sec:SENECA_SENSIM}
\begin{figure*}
    \centering
    \begin{minipage}{0.49\textwidth}
        \centering        \includegraphics[width=\linewidth]{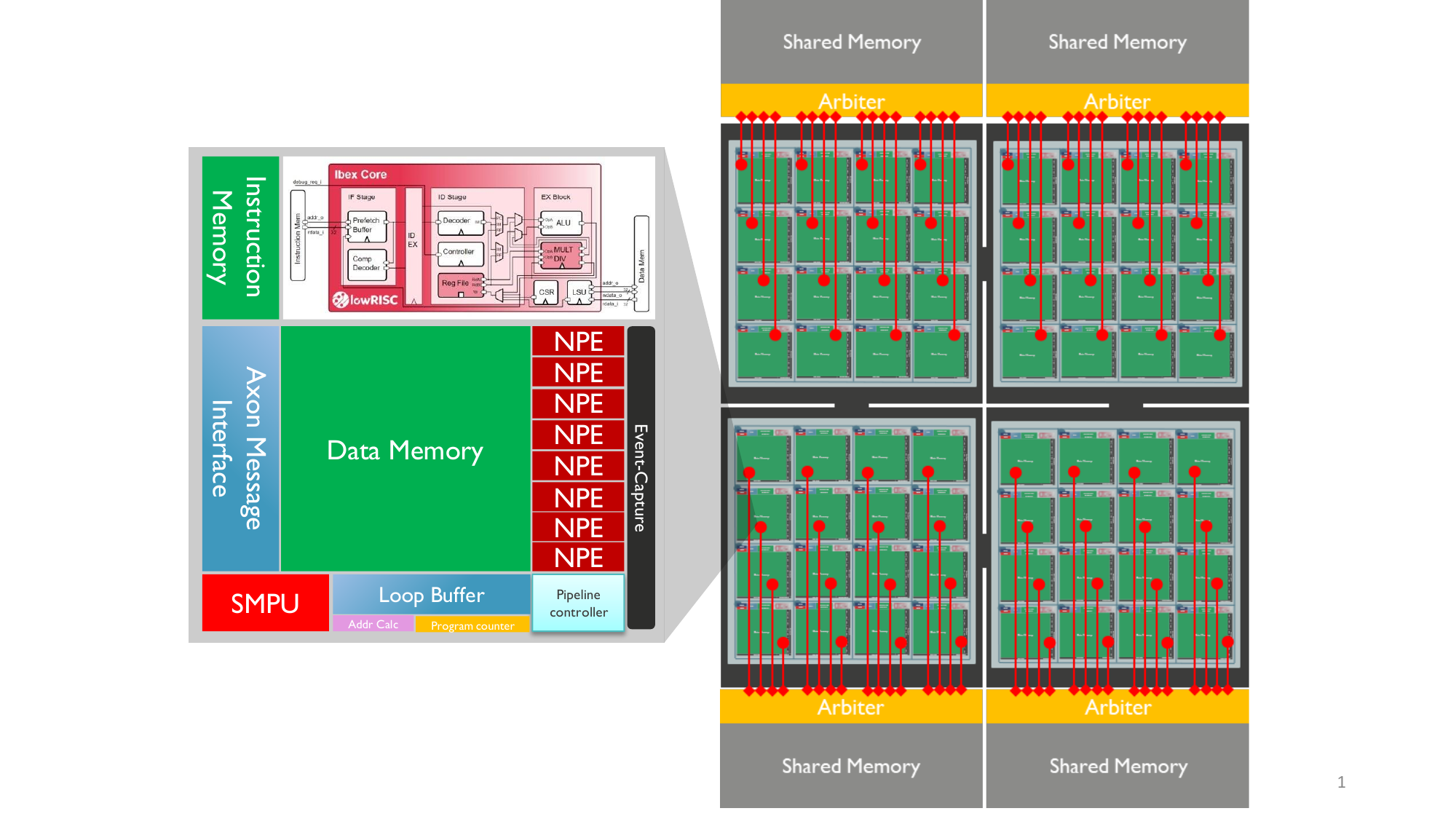}
        \caption{A 64-core SENECA architecture \cite{yousefzadeh_seneca_2022}}
        \label{fig:SENECA_Arch}
    \end{minipage}
    \begin{minipage}{0.49\textwidth}
        \centering
        \vspace{-1.5cm}
        \includegraphics[width=\linewidth]{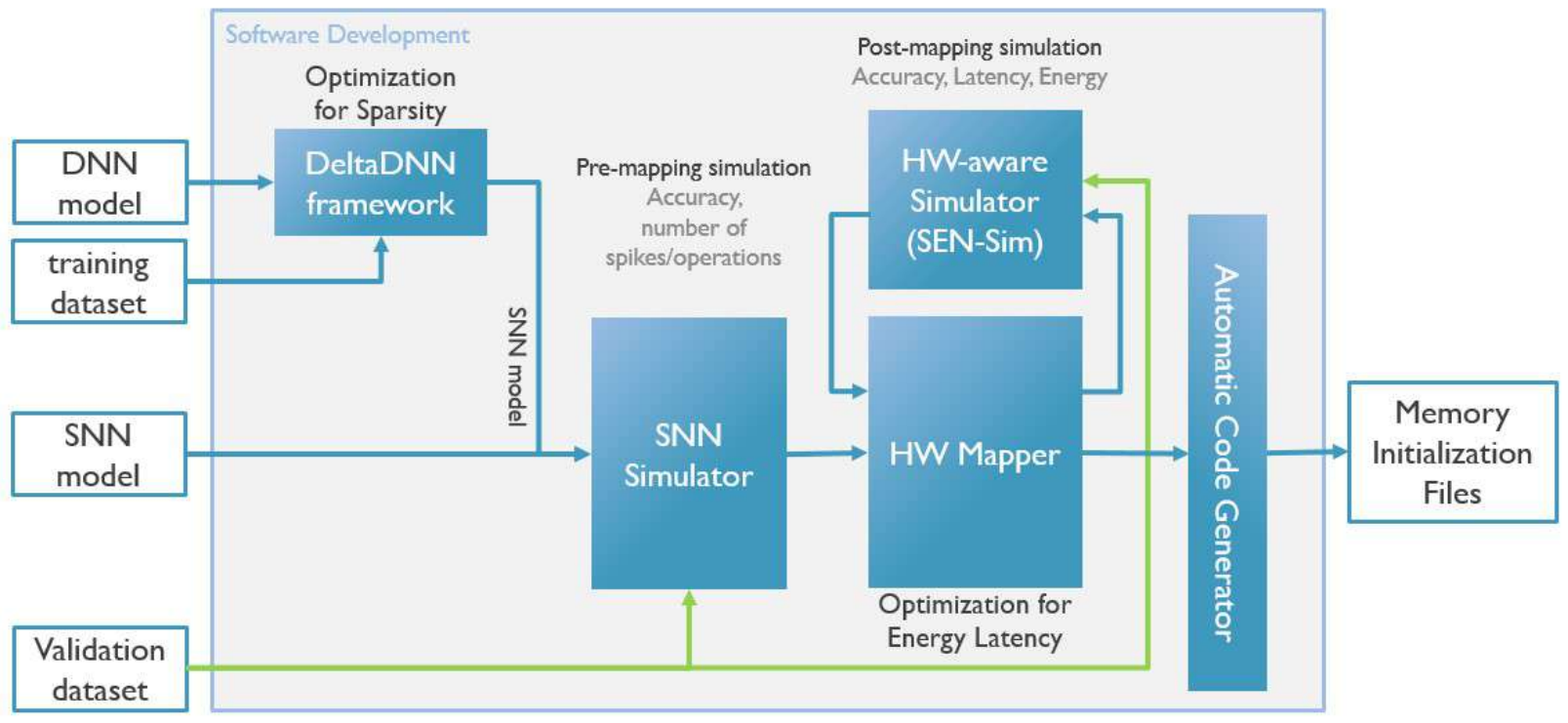}
        \caption{SENECA SDK \cite{yousefzadeh_seneca_2022}}
        \label{fig:SENECA_sdk}
    \end{minipage}
\end{figure*}

SENECA\cite{yousefzadeh_seneca_2022,tang_seneca_2023,tang_open_2023,xu_optimizing_2024} is a RISC-V-based digital neuromorphic processor designed for SNN/ANN computations in edge applications with limited energy. It handles unstructured spatiotemporal sparsity and irregular data transfers with its optimized neuromorphic coprocessor, three-level memory hierarchy, and dual-controlling system. The processor supports data resolutions of 4b, 8b, and Brain-float16, facilitating advanced learning and optimization. SENECA features an event-based NoC that supports multicasting, source-based routing, and data compression. Its configuration is customizable, including the number of cores, the neural processing elements (NPEs) per core, and optional off-chip memory. SENSIM \cite{nembhani_sensim_2024}, an open source event-based simulator developed in Python, helps simulate large SNN/ANN models with hardware-aware parameters, providing estimates of energy and latency. It models systems with data transfer between processors in different clock domains and includes parallel processing support for improved execution time. Unlike other simulators such as PyCarl \cite{balaji_pycarl_2020}, Carlsim \cite{beyeler_carlsim_2015}, and Noxim++ \cite{balaji_mapping_2020}, SENSIM offers detailed temporal energy and latency measurements based on hybrid events and clocks and supports a flexible simulation environment. Fig.\ref{fig:SENECA_Arch} shows an example of SENECA with 64 compute clusters, each comprising an Ibex RISC-V core, memory units, Axon Message Interface (AMI), NoC, Shared Memory Prefetch Unit (SMPU), and NPEs. Fig.\ref{fig:SENECA_sdk} shows the SENECA Software Development Kit (SDK), which includes the DeltaDNN framework, the SNN simulator, the hardware-aware simulator (SENSIM), the SENECA mapper (SENMap) and the code generator. The hardware details for SENECA are detailed in the prior work \cite{xu_optimizing_2024}. The energy, communication, timing, architecture, and simulation parameters extracted from lower-level hardware measurements and used for the experiments are mentioned in the prior work \cite{nembhani_sensim_2024}.

\begin{figure*}
    \centering
    \begin{minipage}{0.49\textwidth}
        \centering        \includegraphics[width=\linewidth]{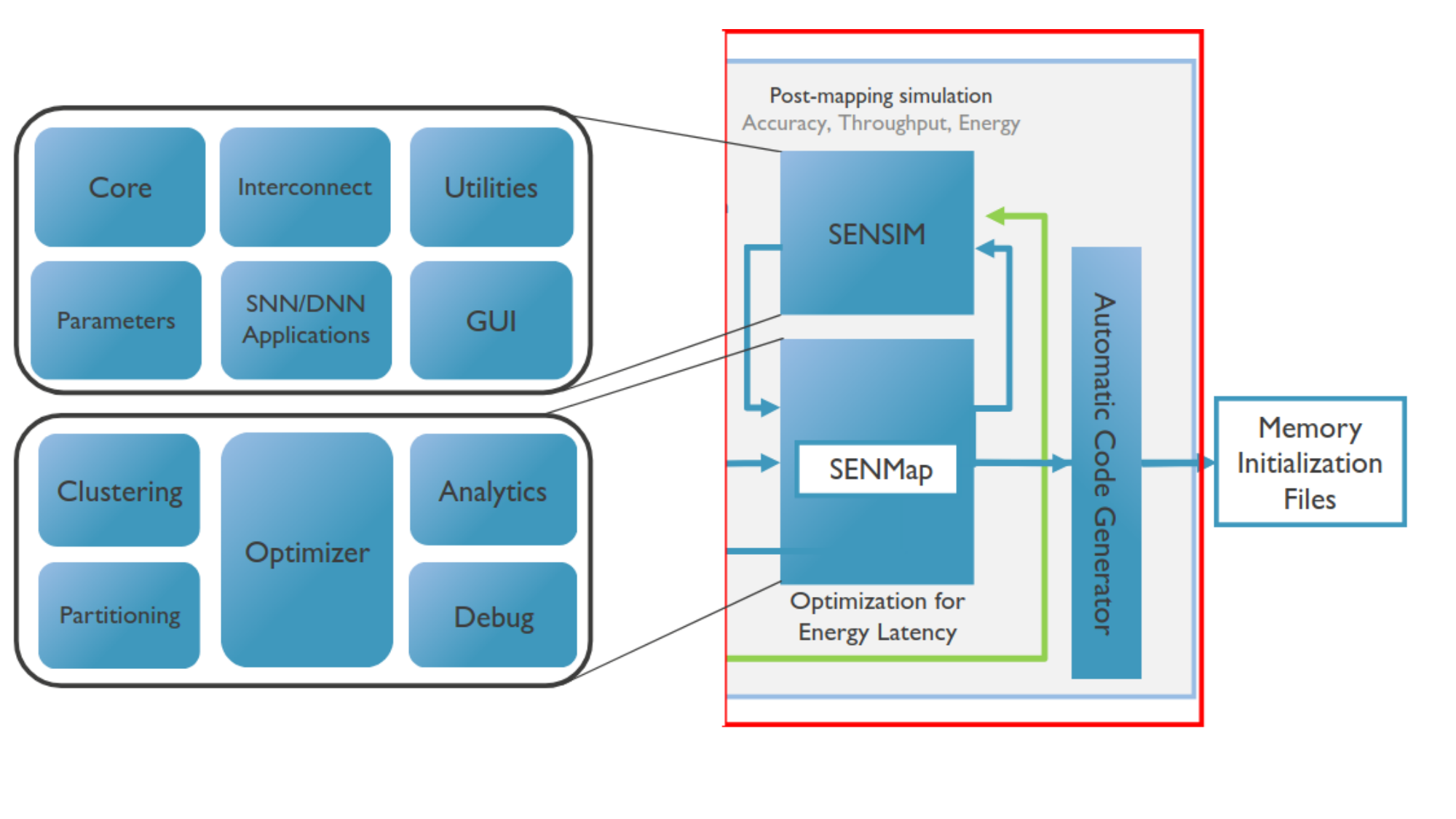}
        \caption{SENSIM and SENMap software framework}
        \label{fig:sensim_senmap}
\end{minipage}
\begin{minipage}{0.49\textwidth}
    \centering
    \includegraphics[scale=0.26]{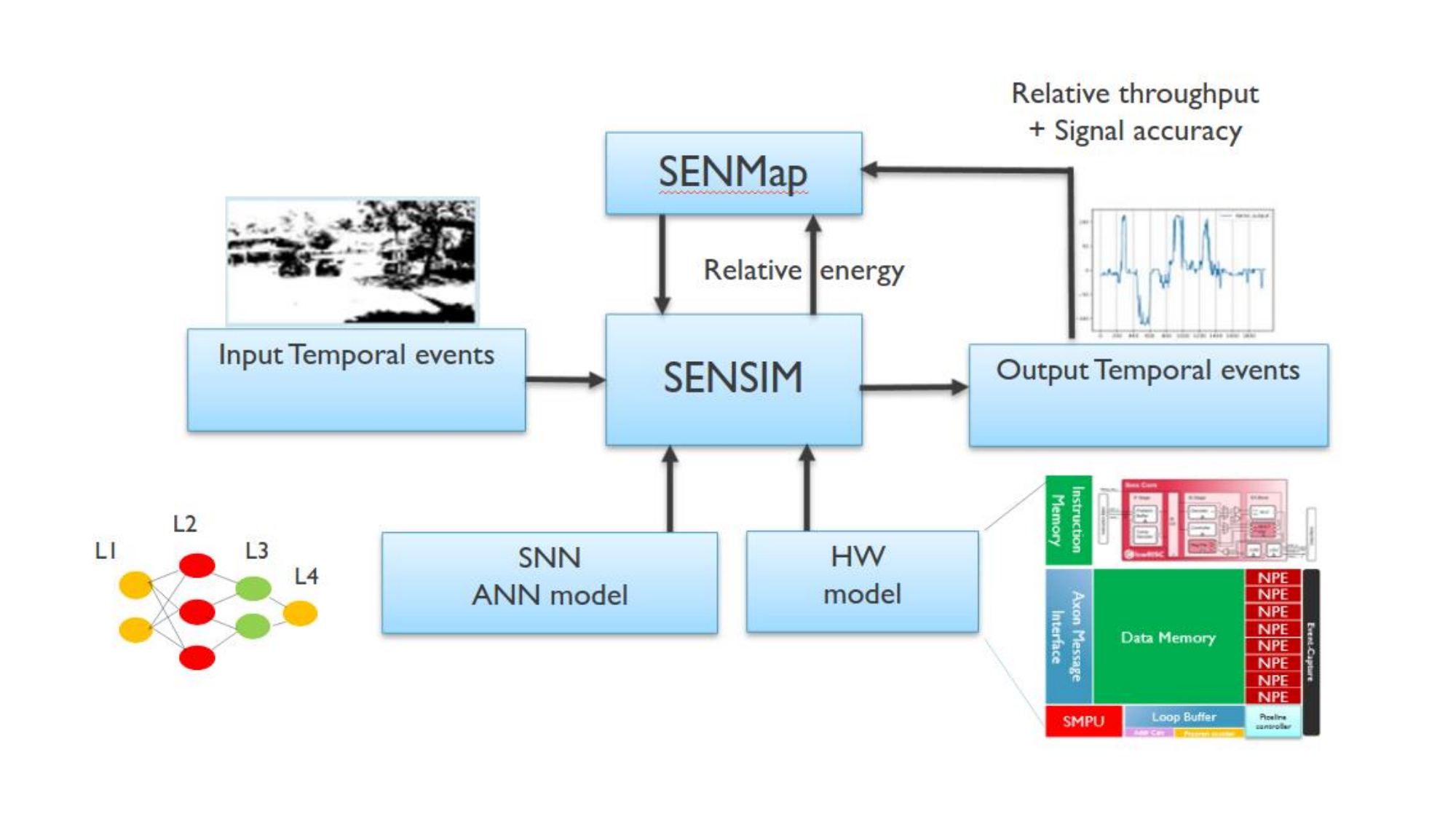}
    \caption{Mapping of PilotNet SNN Models to SENSIM}
    \label{fig:mapping}
\end{minipage}
\end{figure*}
\section{LITERATURE REVIEW}\label{sec:lit_review}
In addition to SENMap, various research efforts have focused on mapping spiking neural networks (SNNs) to neuromorphic systems. The compiler for Loihi \cite{lin_mapping_2018} efficiently maps neurons to Loihi’s cores using a power-efficient greedy algorithm tailored to its crossbar architecture. SpiNeMap \cite{balaji_mapping_2020} organizes SNNs into clusters to minimize spike counts and optimize placement through Particle Swarm Optimization, differing from SENECA’s multicast-based NoC architecture, which employs strategies including inter-spike distortion and energy-efficient hop reduction. SNEAP \cite{li_sneap_2020} uses multilevel graph partitioning to minimize NoC hop counts and enhance energy efficiency, while SpiNNaker2 \cite{kelber_mapping_2020} proposes a similar partitioning strategy to SENECA’s. However, SENMap advances this by incorporating additional architectural parameters. The method in \cite{balaji_run-time_2020} introduces a greedy clustering approach with hill climbing for real-time SNN remapping, though its application to SENECA’s near memory  design presents challenges. The approach in \cite{song_compiling_2020} partitions SNNs to optimize performance but does not address energy consumption, unlike DFSynthesizer \cite{song_dfsynthesizer_2021}, which enhances throughput and energy efficiency through data-flow based scheduling. ALPINE \cite{shreya_mapping_2021} maps various neural network topologies onto crossbar architectures but lacks validation with realistic SNNs, a gap addressed by SENMap through diverse application benchmarking. eSpine \cite{titirsha_endurance-aware_2022} focuses on prolonging memristor lifespan by partitioning workloads and optimizing mapping, contrasting with SENMap’s event-based NoC approach. NeuMap \cite{xiao_optimal_2022} simplifies the mapping process using meta-heuristics and demonstrates significant improvements over previous methods, but focuses on crossbar architectures without considering architectural parameters. The method proposed in \cite{jin_mapping_2023} uses Hilbert curves and force directed algorithms for large-scale SNNs but does not incorporate architectural design space exploration. R-MaS3N \cite{yerima_r-mas3n_2023} improves fault tolerance through neuron reuse and heuristic partitioning, offering scalable solutions for neuromorphic systems. EdgeMap \cite{xue_edgemap_2023} employs multi-objective optimization to enhance performance, latency, and energy efficiency in edge computing scenarios, but SENMap scales more effectively with large neural networks and NoC architectures, estimating a broader set of metrics, including throughput and congestion. Table~\ref{tab:cross_correlation} compares SENMap with other parallel neuromorphic mapping solutions.

\afterpage{
\begin{table*}
\caption{Summary of Parallel Developments in Spiking Neural Network Mapping}
\label{tab:spiking_neural_networks}
\centering
\scriptsize % Reducing the font size (consider \footnotesize after removing resizebox)
% Define a new column type for left-aligned X columns
\newcolumntype{L}{>{\raggedright\arraybackslash}X}
\begin{tabularx}{\textwidth}{@{} l L L L L L L @{}} % l for the first narrow column, L for the rest to wrap
\toprule
\textbf{Mapping} & \textbf{Clustering} & \textbf{Partitioning} & \textbf{Optimizing Parameters} & \textbf{Architecture} & \textbf{Network type} & \textbf{Key difference} \\
\midrule
Loihi compiler \cite{lin_mapping_2018} & greedy, resource-constrained & greedy, custom graph & cross-sectional bandwidth,\newline cut penalty & Loihi, crossbar, in-mem & small/synthetic & novel graph and cut penalty based mapper\\
\addlinespace % Adds a little extra space between rows, from booktabs
SpiNeMap \cite{balaji_mapping_2020} & heuristics (SpiNecluster,\newline spikecount) & metaheuristics (PSO) & spike-count, communication,\newline Inter-Spike Distortion,\newline latency, energy & crossbar,in-mem (DYNAPSE) & small realistic FFN\newline (MLP-MNIST, LeNet-MNIST) & ISI distortion minimization \\
\addlinespace
SNEAP \cite{li_sneap_2020} & multilevel graph & metaheuristics (SA,PSO,Tabu) & energy, spike-count,\newline average hop & crossbar, in-mem & small, synthetic FFN & multilevel graph \\
\addlinespace
SpiNNaker2 \cite{kelber_mapping_2020} & greedy, resource-constrained & layerwise, output channels,\newline height/weight & communication, congestion,\newline datareuse & NoC, NearMem & realistic (ResNet, VGG16) & NearMem mapping,\newline datareuse \\
\addlinespace
Runtime \cite{balaji_run-time_2020} & greedy & hill-climbing & communicated spikes & crossbar, in-mem & small, synthetic & runtime \\
\addlinespace
SNN compiler \cite{song_compiling_2020} & SDFG & greedy & performance, latency & crossbar, in-mem & small, realistic\newline (LeNet, CNN, MLP, Edgdet) & SDFG, Maxplus-Algebra\\
\addlinespace
DFSynthesizer \cite{song_dfsynthesizer_2021} & homogeneous & dataflow scheduling & throughput, energy & crossbar, in-mem & small realistic\newline (VGG16, AlexNet, LeNet) & SDFG \\
\addlinespace
ALPINE \cite{shreya_mapping_2021} & semi-stochastic & Graph-ML, adjacency-list,\newline iterative & throughput & crossbar, in-mem & small, synthetic\newline (FFN, AE, SOM, LSM, RNN) & different SNN typology \\
\addlinespace
eSpine \cite{titirsha_endurance-aware_2022} & KL graph & metaheuristics(PSO) & fault, energy, lifetime & crossbar, memristors NoC & CNN,MLP,RNN & endurance-aware \\
\addlinespace
NeuMap \cite{xiao_optimal_2022} & heuristics (spike-count) & meta-heuristics\newline (multilevel graph partitioning) & spike-firing-rate,\newline communication patterns & crossbar, in-mem & small realistic\newline (CNN-CIFAR, MLP-MNIST,\newline LeNet etc) & multi-level graph partitioning \\
\addlinespace
Very large \cite{jin_mapping_2023} & homogenous & hilbert curve, force-directed & energy, latency, congestion & multicore 2D-NoC & very large realistic/synthetic\newline (AlexNet, MobileNet, ResNet) & very large SNN \\
\addlinespace
R-MaS3N/ NR-NASH \cite{yerima_r-mas3n_2023} & analytical (spike-count,\newline synaptic-connection) & heuristic & accuracy, communication & 3D-NoC & small, synthetic, FFN (MLP) & analytical, 3D-NoC \\
\addlinespace
EdgeMap \cite{xue_edgemap_2023} & custom convex function\newline (size, spike-count) & flow-based,two-stage,\newline multi-objective & energy, communication,\newline throughput, Hop, latency,\newline congestion & agnostic & small realistic\newline (MLP-MNIST, LeNet) & relatively fast \\
\midrule % Use midrule for separation before the highlighted row
\textbf{SENMap (Ours)} & \textbf{homogeneous, greedy} & \textbf{meta-heuristics, multi-objective,\newline channel, height/width} & \textbf{energy, latency, area,\newline accuracy, architectural} & \textbf{scalable, flexible, NoC,\newline NearMem} & \textbf{large realistic SNN/DNN\newline (PilotNet, MLP-MNIST)} & \textbf{rate-based, architectural,\newline closed loop, flexible}\\
\bottomrule
\end{tabularx}
\end{table*}
}

\section{SENMap} \label{sec:senmap}
To optimize processor mapping and configuration for various applications, SENMap (Scalable energy-efficient neuromorphic computing architecture mapper) was developed. It addresses the challenge of mapping event-based large-scale SNN and ANN on SENECA. SENMap mirrors the flexibility of SENECA and SENSIM, offering a range of algorithms and optimization methods, including single and multi-objective strategies. The following subsections detail SENMap's features and improvements. Figure~\ref{fig:mapping} illustrates the mapping of SNN and ANN applications in SENSIM.

\subsection{Clustering and partitioning} \label{subsec:clustering}
\begin{figure}[ht]
    \centering
    \begin{subfigure}
        \centering
        \includegraphics[scale=0.26]{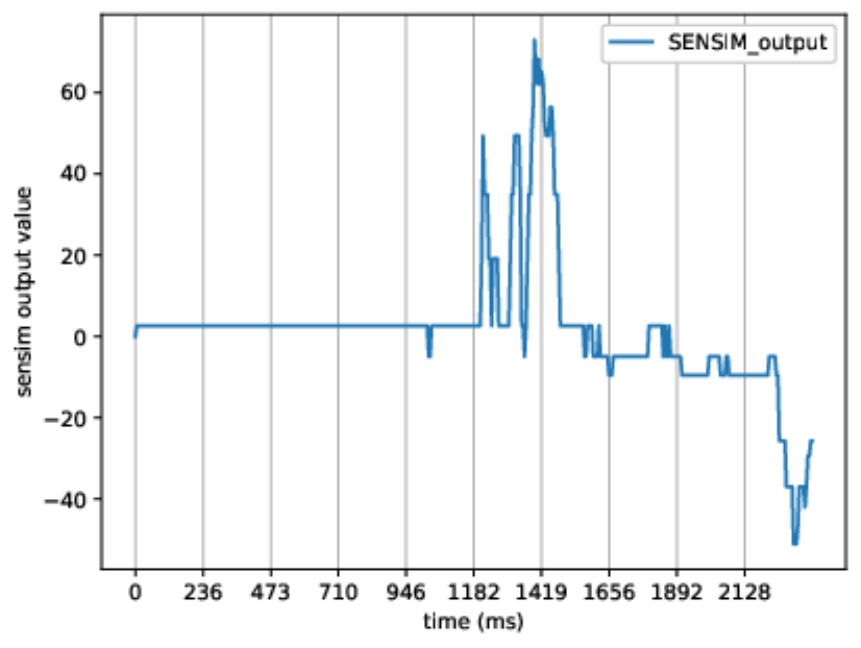}
    \end{subfigure}
    \hfill
    \begin{subfigure}
        \centering
        \includegraphics[scale=0.26]{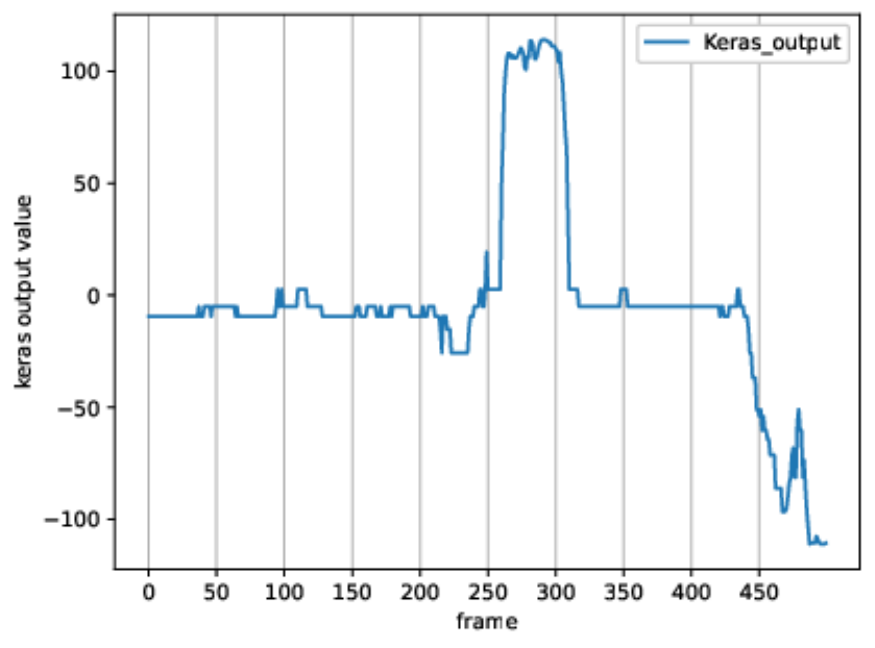}
    \end{subfigure}
    \caption{Output comparison of Inter-spike distorted end signal} \label{fig:sensim_cluster_partition_500}
\end{figure}

SENMap \footnote{Source code of the mapper can be found here. \url{https://github.com/Prithvish04/senMap_paper_submission} and \url{https://github.com/Prithvish04/senMap_experiments}} offers strategies such as partitioning neurons layer-wise, channel-wise, height-wise, and width-wise in a homogeneous and greedy fashion across cores. SENSIM offers clustering layer-wise before distributing them across cores based on the prior information from the analytics framework. In SENSIM, combining layers on a core can lead to interspike distortion and incorrect timestamp updates, leading to erroneous outputs if not handled properly. For example, clustering PilotNet pretrained SNN layers 3 and 4 on the same core can cause deviations, as shown in Figure~\ref{fig:sensim_cluster_partition_500}. For smaller SNN models, interspike distortion is minimized by spike counts, but for large scale we suggest measuring accuracy estimated using end signal correlation methods as detailed in section~\ref{interspike}. 
\subsection{Remapping and compression} 
After partitioning and clustering, the tool includes a mapping compression utility to reduce the design footprint and energy consumption by minimizing throughput across SENECA cores. SENMap employs the following compression schemes:

\begin{itemize}
    \item \textbf{Strict area optimal}: This scheme determines the mesh-size by finding factors of the total number of cores that minimize their sum. For example, with 30 cores, the factor pairs \textbf{[(1,30), (2,15), (3,10), (5,6)]} are evaluated, and the pair \textbf{(5,6)} is selected, resulting in a mesh size of \textbf{[5,6]} or \textbf{[6,5]}.
    \item \textbf{Loose area optimal}: For prime numbers of cores, such as 31, which only has factors \textbf{[(1,31)]}, the layout may be area inefficient. To improve flexibility, an additional core is added to make the total non-prime. For instance, with 31 cores, adding one results in 32 cores with factor pairs \textbf{[(1,32), (2,16), (4,8)]}, where \textbf{(4,8)} is chosen, giving a mesh size of \textbf{[4,8]} or \textbf{[8,4]}. \footnote{Loose area optimal mapping scheme is considered when combining different large neural networks}
    \item \textbf{Strict square}: This scheme aims to create a square-like mesh, even if the number of cores does not perfectly fit into a square. For example, with 26 cores, which typically results in a mesh size of \textbf{[2,13]}, the size is adjusted using a formula to approximate a square shape.  mesh[rows, columns] = $\sqrt{total cores}$
\end{itemize}
\subsection{Bounds and constraints} \footnote{The details on how we use the parameters and form the bounds are mentioned in the previous paper\cite{nembhani_sensim_2024}. All the experiments were carried out in asynchronous time step mode of SENSIM}
Lower bounds are determined by the data memory size, technology node, quantization schemes and the SNN/ANN being mapped. For example, with a 1MB node, M\_{pc} $\leq$ 1MB.
\begin{equation} \label{eq:Memory_cal}
\begin{aligned}
    M_{pc} = 2 \times (N_{npc} + (N_{tpc} \times F_{snn})\times (BW_{states} \\+ BW_{outputs})) + (N_{wpc} + N_{bpc}) \times BW_{weights}
\end{aligned}
\end{equation}
where: 
\begin{table}[H]
    \centering
    \begin{tabular}{ll}
        $N_{npc}$ & neurons per core \\
        $N_{wpc}$ & weights per core \\
        $N_{bpc}$ & biases per core \\
        $N_{tpc}$ & thresholds per core \\
        $M_{pc}$  & memory per core \\
        $BW_{states}$ & bit-width of neuron states \\
        $BW_{outputs}$ & bit-width of outputs \\
        $BW_{weights}$ & bit-width of weights and biases \\
        $F_{snn}$ & flag for SNN (true if SNN) \\
    \end{tabular}
\end{table}
With SENSIM, the chip architect determines the upper bound, as the total number of cores is considered infinite.
\afterpage{
\begin{figure*}
    \centering
    \begin{minipage}{0.99\textwidth}
    \framebox[\textwidth]{
    \begin{minipage}{\textwidth}{
        \begin{subfigure}
            \centering            \includegraphics[width=0.34\textwidth]{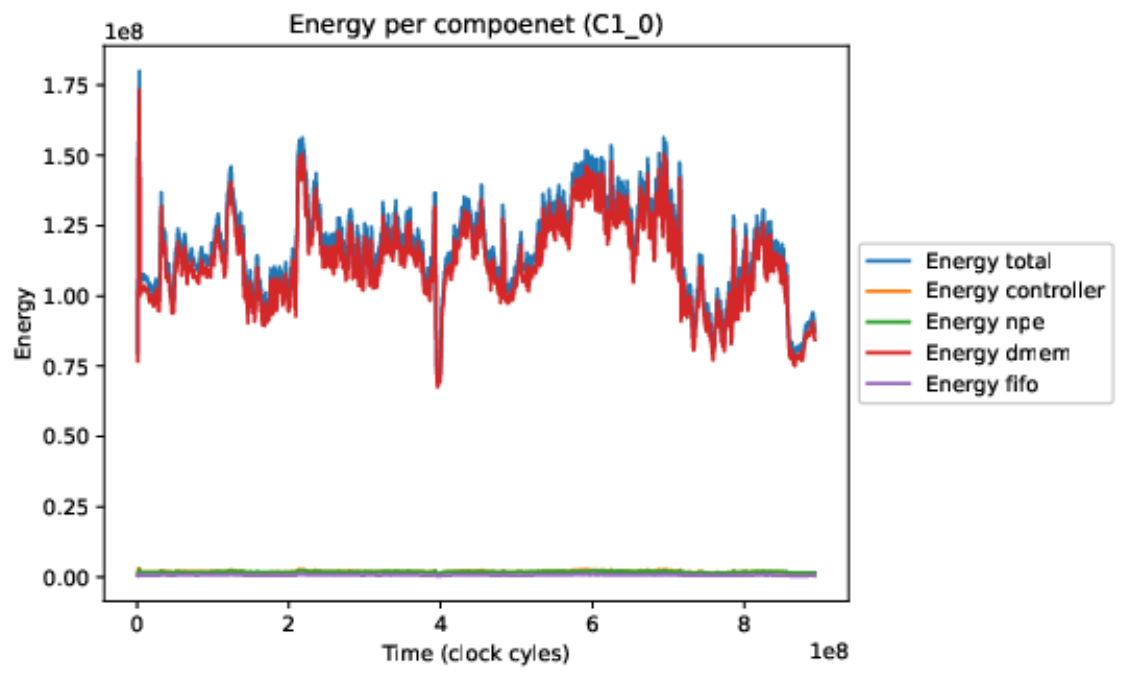}
            \label{subfig:C0_0}
        \end{subfigure}
        % \begin{subfigure}
        %     \centering            \includegraphics[width=0.24\textwidth]{figures/analysis/Peak_in_out_queue_C8_0.pdf}
        %     \label{subfig:C1_0}
        % \end{subfigure}
        \begin{subfigure}
            \centering
            \includegraphics[width=0.30\textwidth]{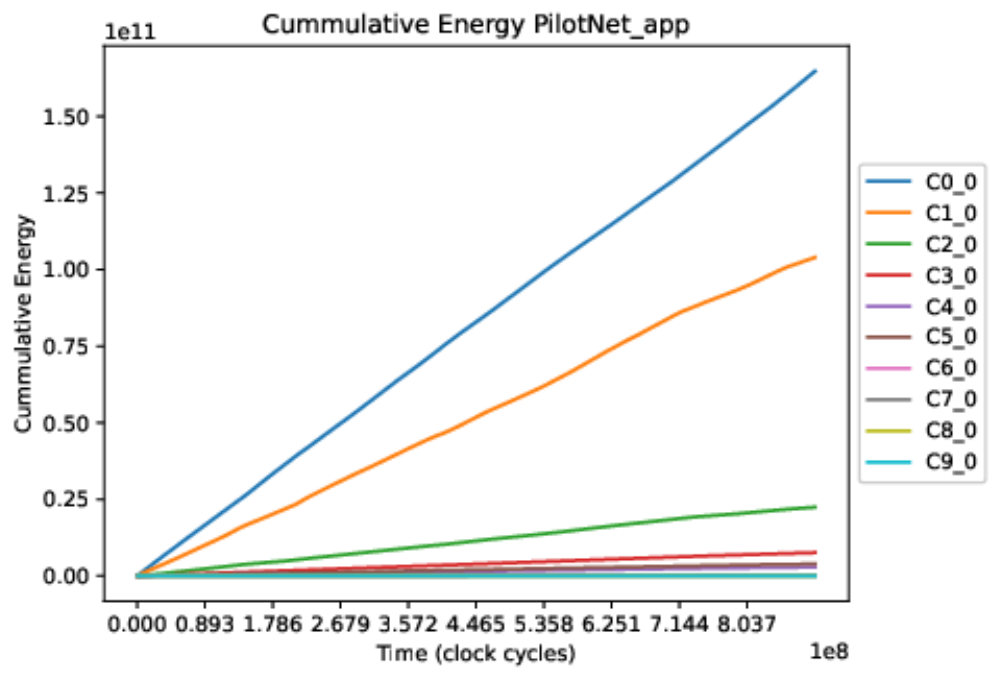}
            \label{subfig:C0_1}
        \end{subfigure}%
        \hfill
        \begin{subfigure}
            \centering
            \includegraphics[width=0.30\textwidth]{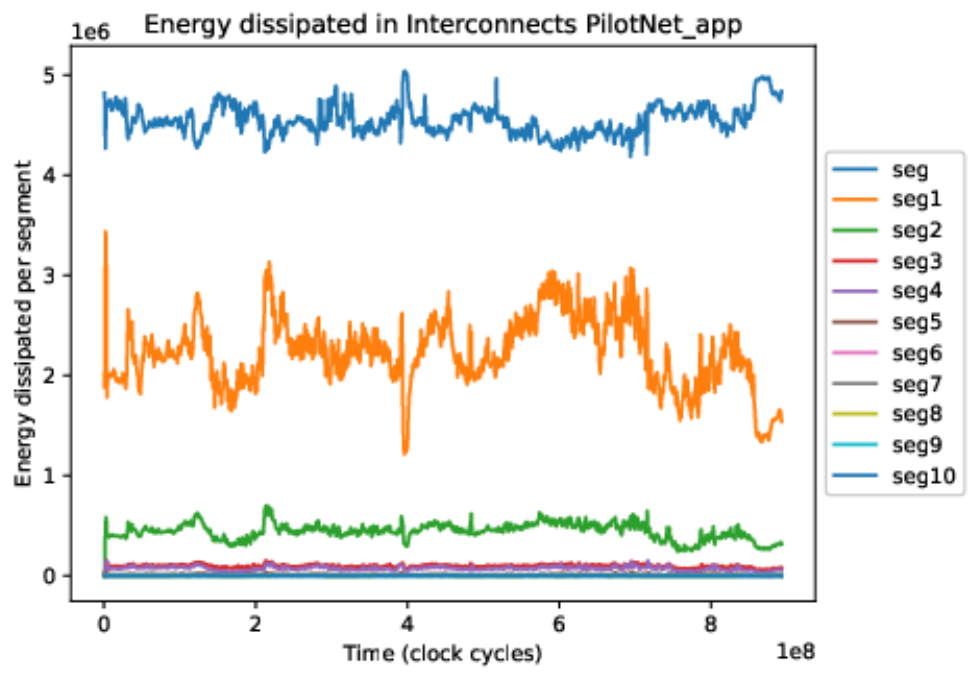}
            \label{subfig:C1_1}
        \end{subfigure}%
        }
        \end{minipage}
    }
    \framebox[0.6\textwidth]{
    \begin{minipage}{0.6\textwidth}
        \begin{subfigure}
            \centering            \includegraphics[width=0.48\textwidth, height=6cm]{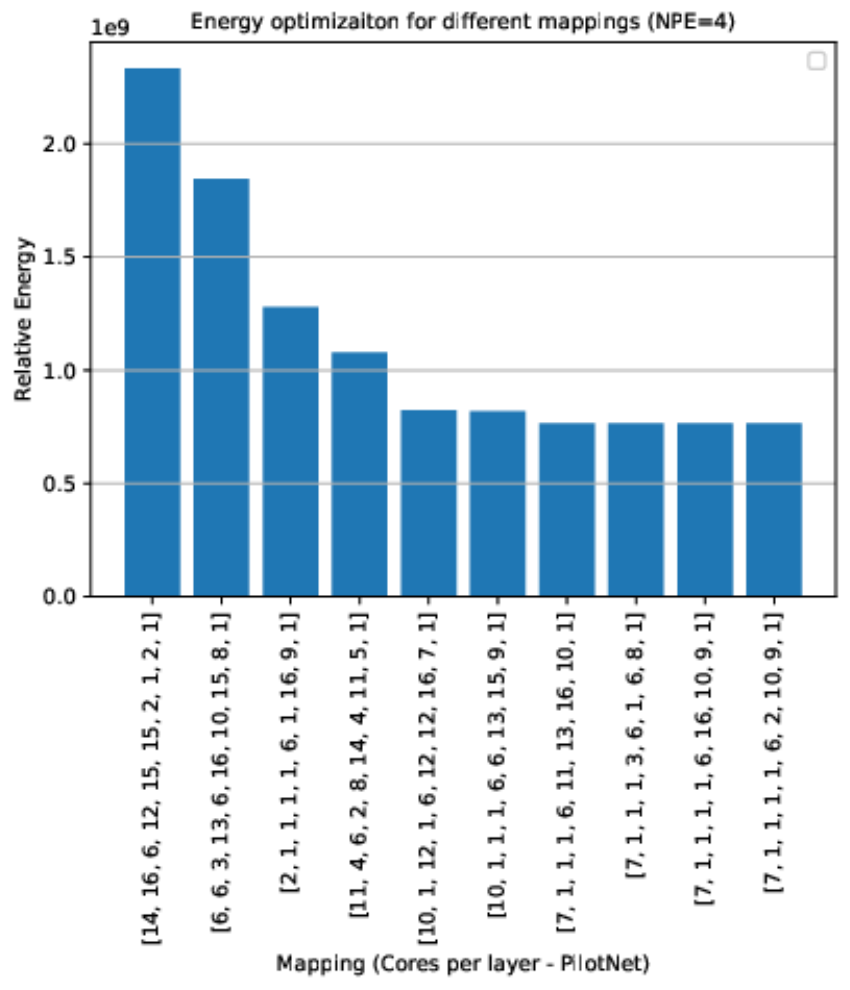}
            \label{subfig:C0_3}
        \end{subfigure}
        \begin{subfigure}
            \centering            \includegraphics[width=0.48\textwidth, height=6cm]{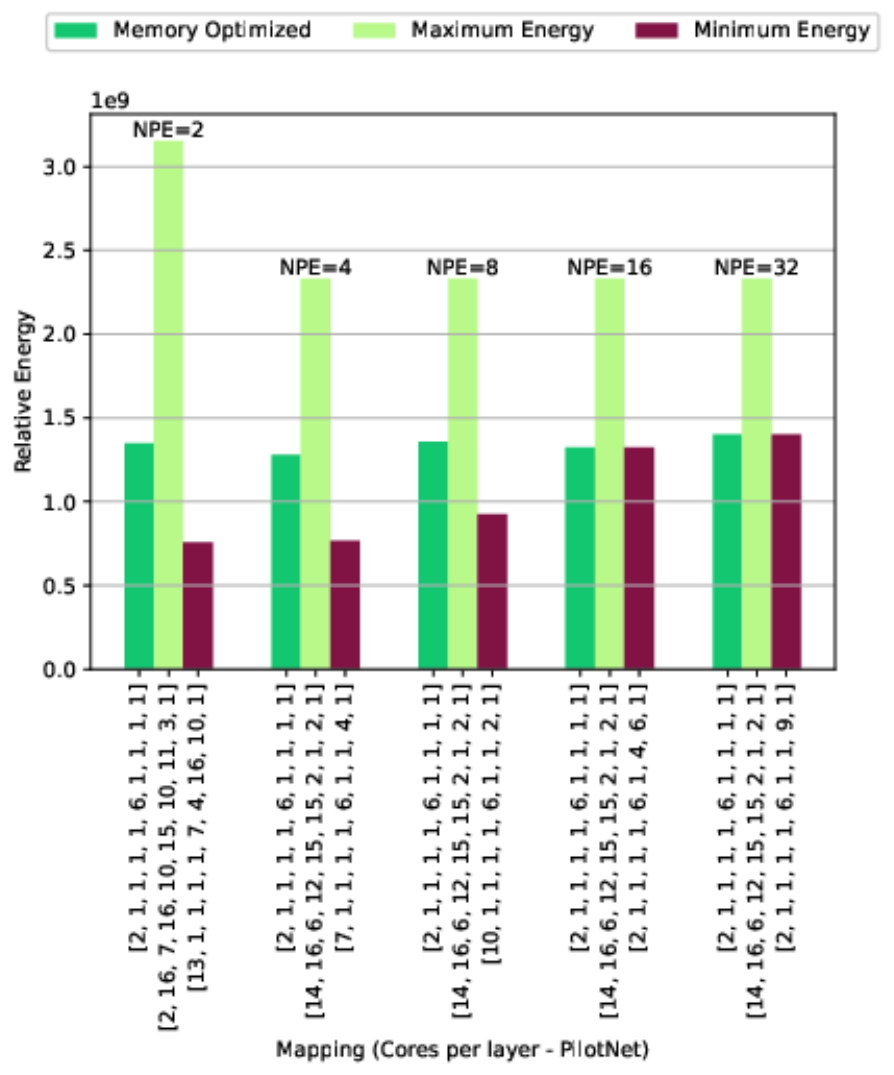}
            \label{subfig:C1_3}
        \end{subfigure}%
    \end{minipage}
}\framebox[0.4\textwidth]{%
\begin{minipage}{0.39\textwidth} 
\tiny
\setlength{\DTbaselineskip}{6pt}
\dirtree{%
.0 experiments/.
    .2 Mnist\_app/.
    .2 PilotNet\_app/.
        .3 BRKGA\_0\_2022\_07\_23\_23:22:35/.
        .3 NSGA2\_1\_2022\_08\_05\_14:20:58/.
            .4 Energy/.
                .4 2022\_08\_05\_14:21:44\_14047.
                    .6 gui\_setting.csv.
                    .6 output\_snapshot.csv.
                    .6 snapshots\_cores.csv.
                    .6 snapshots\_interconnects.csv.
                    .6 summary.txt.
                .5 2022\_08\_05\_14:21:44.
            .4 Latency/.
                .5 2022\_08\_05\_14:21:44\_14047.
                    .6 gui\_setting.csv.
                    .6 output\_snapshot.csv.
                    .6 snapshots\_cores.csv.
                    .6 snapshots\_interconnects.csv.
                    .6 summary.txt.
            .4 NSGA2\_sum\_2022\_08\_05\_14:20:58/.
                .5 algo.prm.
                .5 latOpt.csv.
                .5 energyOpt.csv
                .5 sim.prm.
                .5 plot\_cores.
                .5 plot\_interconnect.
        .3 PSO\_0\_2022\_08\_01\_05:28:39.
            .4 2022\_08\_01\_06:59:13\_140477164676764.
            .4 PSO\_summary\_2022\_08\_01\_05:28:39.
}
\end{minipage}
}
\caption{Analytics and data collection framework for Pre and Post optimization}
\label{fig:analytics}
\end{minipage}
\end{figure*}}
\subsection{Optimization algorithms and problem design} SENMap utilizes single, multi-, and many-objective optimization through the Pymoo framework \cite{blank_pymoo_2020}, which includes algorithms such as Genetic Algorithm (GA), Particle Swarm Optimization (PSO), ISRES \cite{tp_runarsson_search_2005}, NSGA-II, NSGA-III \cite{vesikar_reference_2018}, and MOEA\/D \cite{zhang_moead_nodate}. Integrating PyMOO allows SENMap to leverage various optimization algorithms and parameters, which is beneficial for tackling the nonlinear problems involved in mapping \& closed-loop synthesis of large neural networks. Metaheuristics in particular are the most suitable for these challenges considering the diversity of SNNs, the temporal dynamics in the system, and the evolving nature of SENECA resulting in nonlinearities in the system at scale. However, hyperparameter tuning or heuristics reduces complexity and solution time. To formulate the problem for optimization, SENMap evaluated several metaheuristics and tools, including jMetalPy \cite{benitez-hidalgo_jmetalpy_2019}, PyGMO \cite{biscani_parallel_2020}, Open Beagle \cite{gagne_open_2002} and Opt4J \cite{lukasiewycz_opt4j_2011} PyMOO was the most suitable, supporting a wide range of optimization problems, including random search for binary variables, discrete variables, permutations, mixed variable types, custom variables, biased initialization, and subset selection. 

\subsection{Inter-spike distortion}\label{interspike} 
Inter-spike distortion is a significant challenge when mapping neurons onto neuromorphic hardware \cite{balaji_mapping_2020}. While previous approaches \cite{balaji_pycarl_2020,beyeler_carlsim_2015} focus on minimizing distortion by analyzing individual spike timing, our method assesses signal integrity by comparing the correctness of two end signals, $x(t)$ and $y(t)$, using cross-correlation to produce a normalized correlated signal $z(t)$. The approximate shift in the end signal is then estimated.
\begin{equation} \label{eq:combined}
z(t) = \frac{\text{corr}(x(t),y(t))}{\sqrt{(x(t)\cdot y(t))(x(t)\cdot y(t))}};t_{shift} = \underset{t}{\arg\max}\,z(t)
\end{equation}
\subsection{Data collection and visual debugging} 
SENMap collects data for all the optimization iterations to train a model, approximate the energy and latency while running optimization experiments. The data could be used towards designing data-driven heuristics. Adaptive mapping in real time and energy reduction is a feature that is in development~\cite{prithvish_efficient_nodate}.
\subsection{Flexible SNN\/ANN replacement} Since SENSIM takes into account the mapping of several large-scale SNN and DNN with temporal delta activation convolution and dense layers, hence SENMap was designed in such a way that easy selection and replacement of SNN/ANN is possible. SENMap gives the flexibility to optimize on a hybrid SNN-ANN architecture \cite{noauthor_towards_nodate, ahmed_hybrid_2024} in the future and other topologies as suggested in \cite{shreya_mapping_2021}. Figure~\ref{fig:analytics} gives an overview of details included in the data collection and debugging framework. The framework also suggests that energy in the core is accumulating linearly per core when PilotNet is mapped to it. On varying parameters, the accumulation of the energy per core is piecewise linear.\footnote{The SNNs used in the experiments are not trained from scratch but are rather converted from pretrained ANN networks and spatial temporal sparsity is extracted from them}
\subsection{Towards closed loop synthesis} SENMap is not only designed for optimal neural-network mappings on SENECA but also supports hardware architecture and SNN/ANN event model parameters co-optimization to optimize energy, latency, area, and accuracy. It maps neuron states, weights, biases, and thresholds to the core while allowing flexibility in neuromorphic parameters like processing elements, memory, bit-width, clock frequency, and flit-width.
\subsection{Parallel metaheuristics} Parallel meta-heuristics enhance solution quality by running multiple searches simultaneously, particularly in population-based algorithms such as PSO \cite{kennedy_particle_1995}. The Pymoo framework leverages Python's multiprocessing and Dask \cite{rocklin_dask_2015} to accelerate tasks including processing 2000 images, which can be substantially improved with parallel execution on a 30-core cluster node, requiring around $\approx$ 6 GB of memory per thread.

\section{Experimental Results} \label{sec:experiments}

\begin{figure*}[ht]
    \centering
    \begin{minipage}{\textwidth}
     \begin{subfigure}
        \centering        \includegraphics[width=0.185\textwidth]{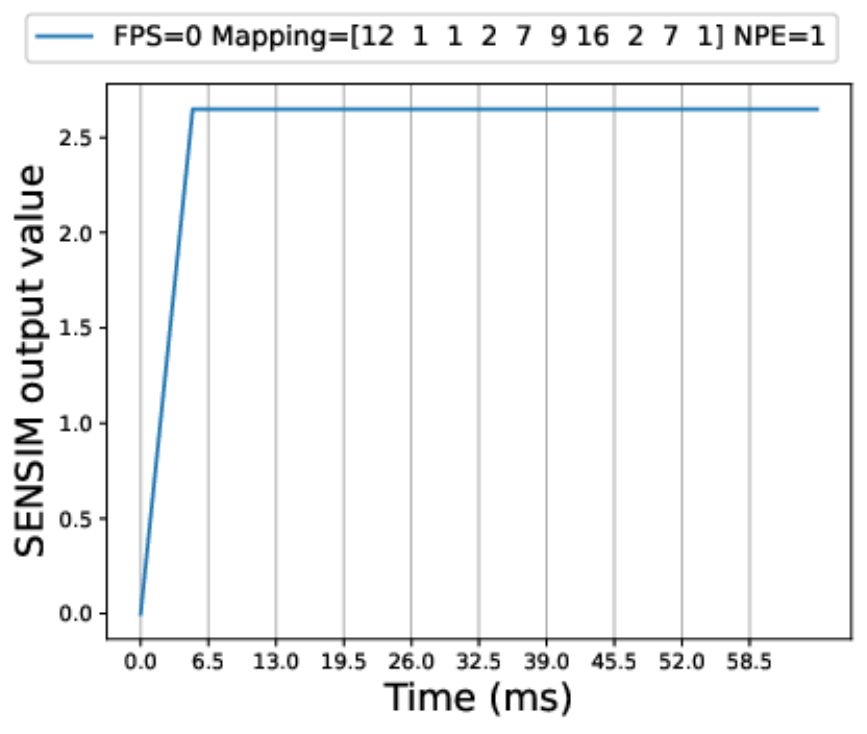}
     \end{subfigure}
     \begin{subfigure}
         \centering        \includegraphics[width=0.185\textwidth]{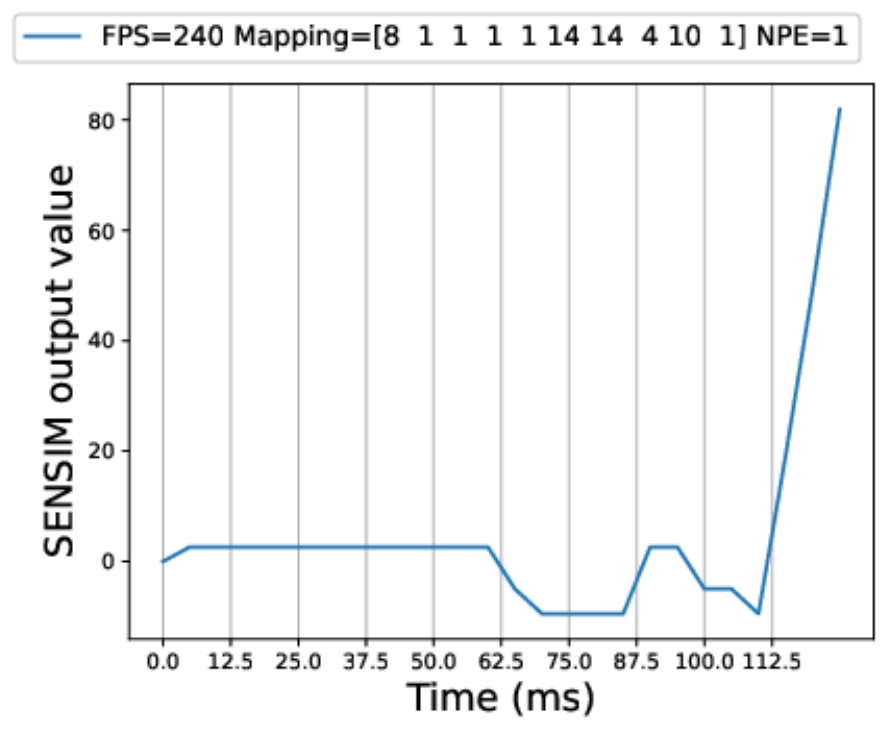}
     \end{subfigure}
    \begin{subfigure}
         \centering         \includegraphics[width=0.185\textwidth]{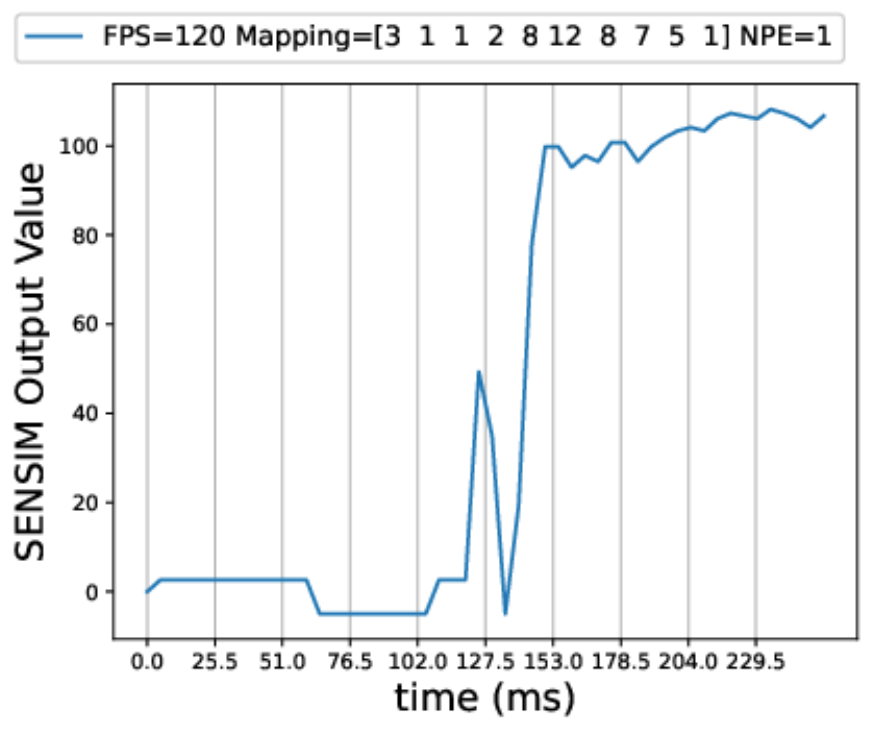}
     \end{subfigure}
     \begin{subfigure}
         \centering         \includegraphics[width=0.185\textwidth]{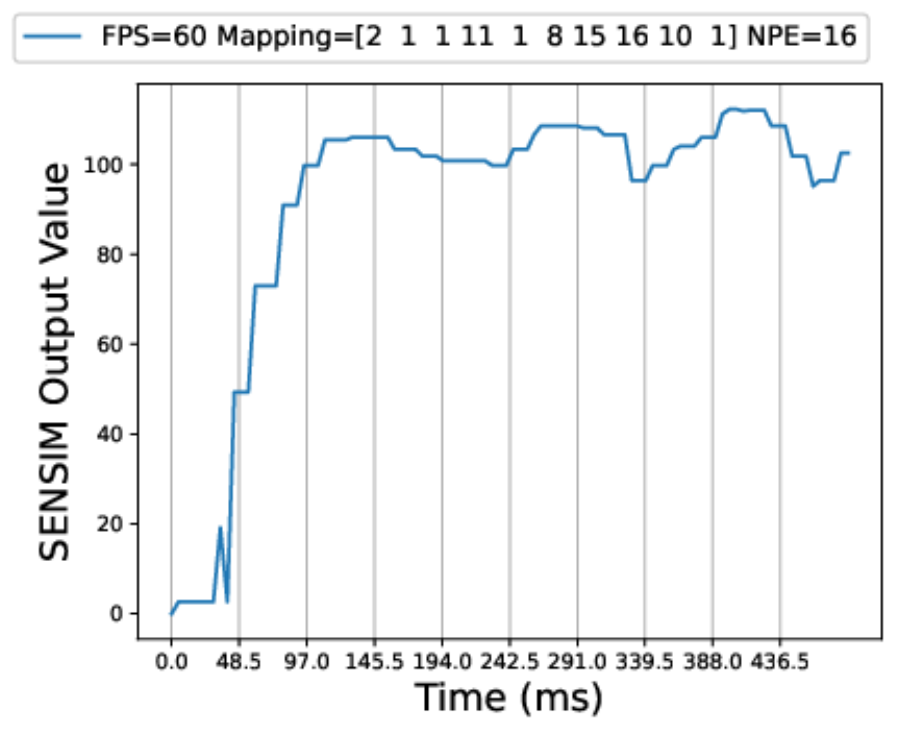}
     \end{subfigure}
     \begin{subfigure}
         \centering         \includegraphics[width=0.185\textwidth]{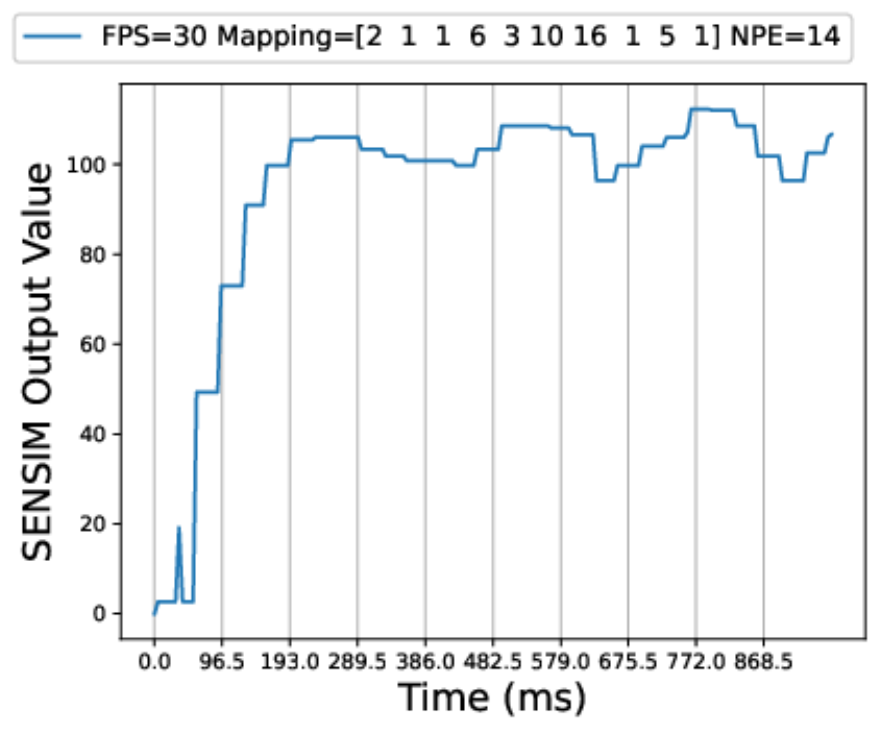}
    \end{subfigure}\caption{SENSIM output comparison (\#260-\#290) frames for different mapping schemes, event/frame rate \& neural processing elements}
    \label{fig:sensim_output_comparison}
     \end{minipage}
\end{figure*}
\begin{figure*}[ht]
    \centering
    \begin{minipage}{\textwidth}
     \begin{subfigure}
         \centering  \includegraphics[width=0.184\textwidth]{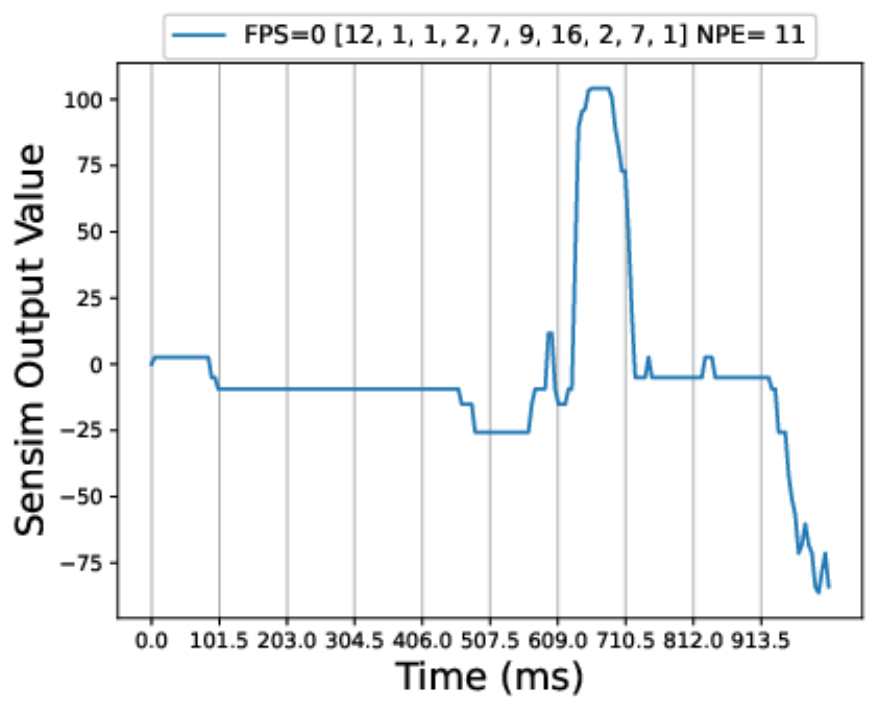}
     \end{subfigure}
     \begin{subfigure}
         \centering         \includegraphics[width=0.184\textwidth]{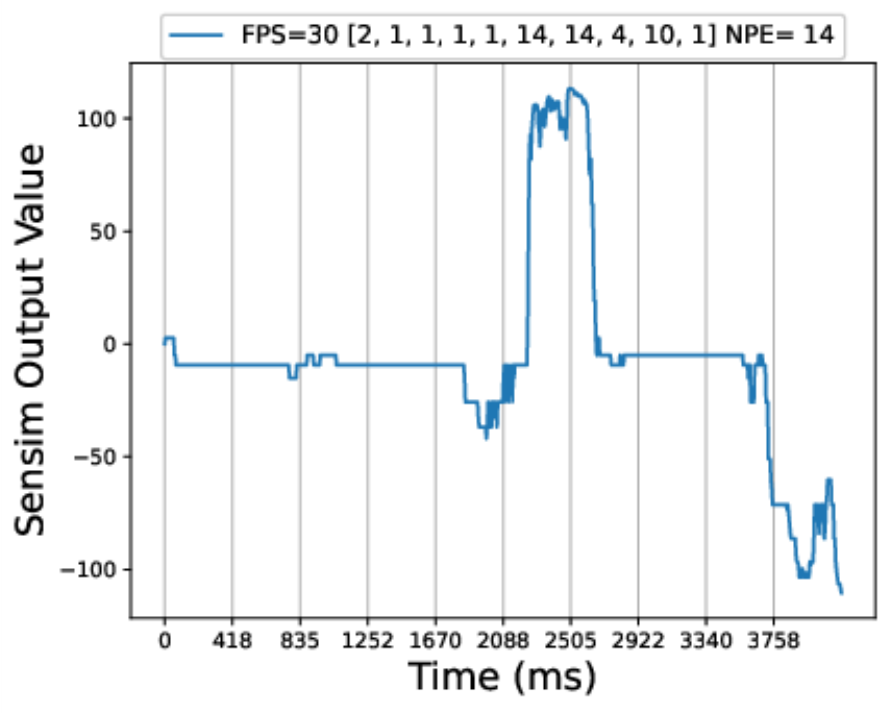}
     \end{subfigure}
     \begin{subfigure}
         \centering         \includegraphics[width=0.184\textwidth]{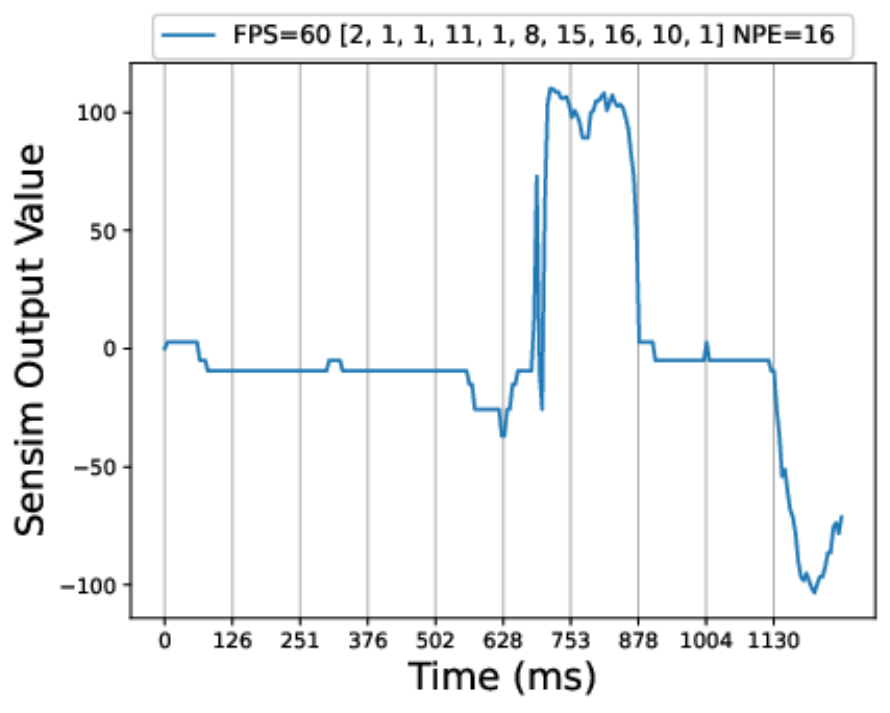}
     \end{subfigure}
     \begin{subfigure}
         \centering                \includegraphics[width=0.184\textwidth]{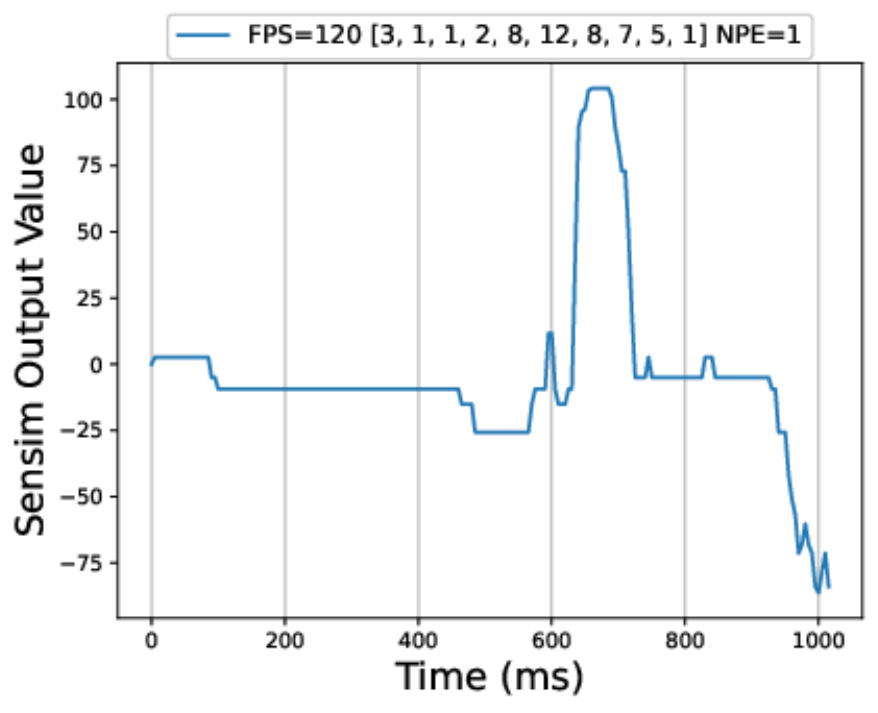}
     \end{subfigure}
     \begin{subfigure}
         \centering               \includegraphics[width=0.184\textwidth]{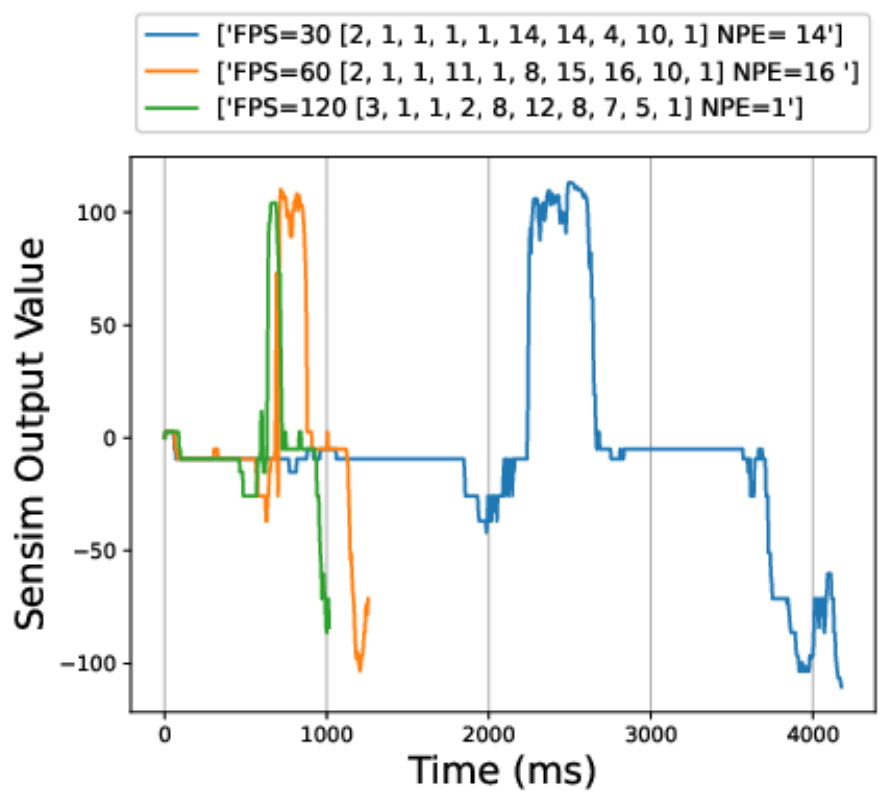}
     \end{subfigure}
    \caption{Comparing the output for fps=30,60,120,0 (event-based) for(\#0-\#500) frames for different mapping \& number of NPEs}\label{fig:30_60_120_fps_pilotnet_comparison}
    \end{minipage}  
\end{figure*}
\begin{figure}[ht]
    \centering \includegraphics[width=0.40\textwidth]{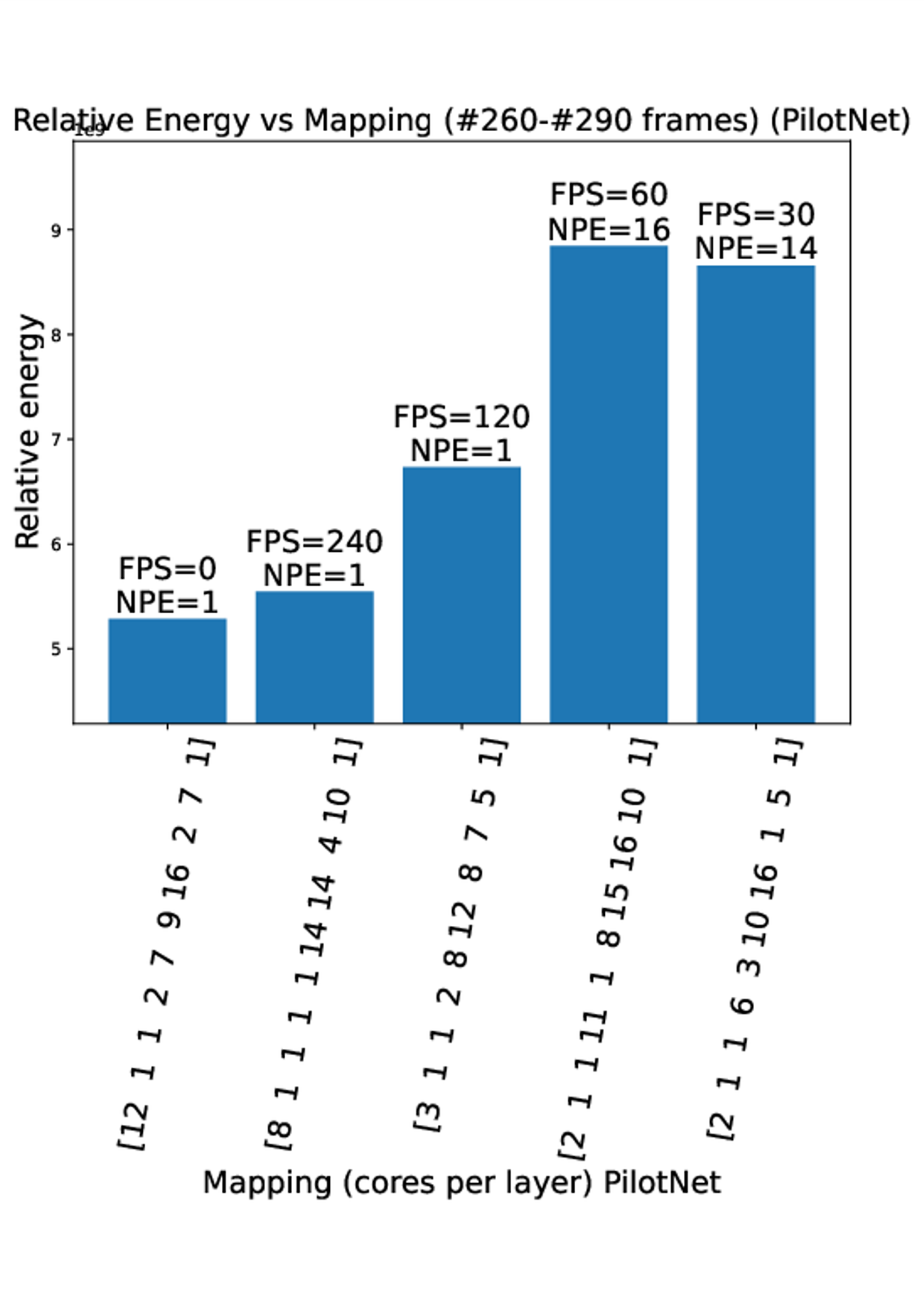}
    \caption{Relative energy vs mapping for (\#260-\#290) frames (PilotNet) at different rates}\label{fig:core_per_layer_vs_mapping_diff_rates}
\end{figure}

This section presents experiments with SENMap, focusing on optimizing architectural and event rate-based parameters for the neural network \cite{bojarski_end_2016} (subsequently named PilotNet). Using varied dataset sizes (\#10-\#30 frames), we observed that dataset size and energy efficiency minimally affected results. A mathematical model for the neuromorphic core, detailed in \cite{nembhani_sensim_2024}, revealed a nonlinear system due to operations like \textbf{max}, necessitating adaptable meta-heuristics over fixed heuristics for dynamic demands in ANN and SNN applications. Despite the variety of meta-heuristics available, no single approach universally excels due to the no-free lunch theorem \cite{ho_simple_2002}; effectiveness depends on the problem structure, hence for experiments and synthesis we used GA and NSGA-2 with hyper-parameters mentioned in Table~\ref{tab:combined_params}, as implemented in PyMOO, to balance multiple objectives like energy, latency, and accuracy.

Parallel meta-heuristics reduced time-to-solution from 3 days to 3 hours. The results indicate that frame rate impacts the precision of PilotNet inference and mapping efficiency, as illustrated in Figures~\ref{fig:sensim_output_comparison},\ref{fig:30_60_120_fps_pilotnet_comparison} \& \ref{fig:core_per_layer_vs_mapping_diff_rates}. These figures show that increasing frame rates, particularly to 120 fps, affects mapping, making asynchronous transfer ideal for SNNs, where fewer NPEs require additional cores for SIMD arrays. We measure a 40\% energy improvement from right to left in Figure~\ref{fig:core_per_layer_vs_mapping_diff_rates} as we increase the rate and an optimal mapping requiring just 1 NPE and an increase in cores. In ANNs, rates beyond a threshold can lead to signal interleaving, affecting output accuracy. SENECA's flexibility as an edge AI chip supports both SNN and ANN needs, with SENMap providing optimal mappings across applications. The correlation values for the SENSIM output signals for the PilotNet dataset as shown in Figure~\ref{fig:core_per_layer_vs_mapping_diff_rates} and the values can also be compared here in Table~\ref{tab:cross_correlation}. The end signals are normalized before correlation and the time shift is estimated in millisecond after scaling up with the time-step in the simulation used.\footnote{fps=0 corresponds to a complete event driven system}

\begin{table}[htbp]
    \caption{Algorithm Parameters for GA and NSGA2}
    \label{tab:combined_params}
    \centering
    \begin{tabular}
    {p{0.25\linewidth} p{0.3\linewidth} p{0.3\linewidth}} 
    \hline 
    \textbf{algo param} & \textbf{GA value} & \textbf{NSGA2 value} \\ \hline 
    population size & 30 & 40 \\
    mutation type & integer polynomial mutation (eta=3.0) & integer polynomial mutation (eta=3.0) \\
    crossover type & integer SBX (eta=3.0) & integer SBX (eta=3.0) \\
    sampling type & integer (random selection) & integer (random selection) \\
    off-springs & - & 10 \\ \hline % 
    \end{tabular}
\end{table}

\begin{table}[htbp]
\centering
\caption{Signal Cross-Correlation Results}
\label{tab:cross_correlation}
\begin{tabular}{ccc}
\hline
\textbf{Correlation Pair} & \textbf{Correlation value} & \textbf{Time shift(ms)} \\
\hline
$\text{corr}(fps=30, fps=60)$ & 0.89 & 1565 \\
$\text{corr}(fps=60, fps=120)$ & 0.88 & 65 \\
$\text{corr}(fps=30, fps=120)$ & 0.90 & 1705 \\
$\text{corr}(fps=0, fps=120)$ & 1 & 0 \\
$\text{corr}(fps=0, fps=60)$ & 0.88 & 65 \\
$\text{corr}(fps=0, fps=30)$ & 0.90 & 1705 \\
\hline
\end{tabular}
\end{table}

\section{Discussion} \label{discussion}
Over the last four years, SENECA has undergone several changes. Initially, it featured a SIMD NPE with a loop buffer \cite{yousefzadeh_seneca_2022}, emulating a GALS quantized system. It has since evolved into a double-controlled system with varying levels of synchronicity in a core-to-core configuration \cite{xu_optimizing_2024}.  With these varying changes, apping large pretrained SNNs and ANNs is expected to grow in importance.
Event rate as a parameter not identified in prior work is something that needs to be considered when taking the mapping into account for event-based systems. We conclude that the event rate is a critical parameter in scalable, energy-efficient neuromorphic systems. This raises further questions: At what rate were frames acquired? What event rates were models trained on? What sensors were used to train these existing models?

\section{Conclusion and Future Work} \label{sec:conclusion_future_work}
This paper addressed key challenges in mapping large SNNs onto hardware platforms through SENMap, a versatile framework integrated with existing software frameworks. We presented flexible heuristic and meta-heuristic algorithms, including genetic algorithms and particle swarm optimization, to improve neuron partitioning and mapping efficiency. Addressing inter-spike distortion and clustering challenges, our approach minimized latency and energy while managing constraints including chip area limitations for large SNN applications. SENMap achieved around 40\% energy efficiency without compromising accuracy. Future work could enhance SENMap by allowing layer combinations to reduce signal distortion, utilizing 3D core stacking and 2.5D and 3D memory configurations. \cite{zhang_challenges_2022} to improve large-scale SNN mapping, and expanding synthesis to include communication and routing parameters. Multi-objective optimization could further refine chip synthesis, while scaling SENMap for larger applications such as ResNet \cite{he_deep_2015} would improve real-time performance and energy efficiency, advancing algorithm-hardware-software co-optimization for hybrid SNN and ANN applications.

\section{ACKNOWLEDGMENT}\label{sec:acknwledgements}
This work was partially funded by research and innovation projects REBECCA (KDT JU under grant agreement No. 101097224), NeuroKIT2E (KDT JU under grant agreement No. 101112268), and NimbleAI (Horizon EU under grant agreement 101070679), and fully supported by friends and family.

\bibliographystyle{ieeetr}
\bibliography{references}

\end{document}